# Redundancy Resolution in Kinematic Control of Serial Manipulators in Multi-Obstacle Environment

Wanda Zhao[1], Anatol Pashkevich[1,2] and Damien Chablat[1,3]

[1] Laboratoire des Sciences du Numérique de Nantes(LS2N), UMR CNRS 6004, Nantes, France
[2] IMT Atlantique Bretagne Pays de la Loire, Nantes, France
[3] Centre National de la Recherche Scientifique (CNRS) , France
Wanda.Zhao@ls2n.fr; Anatol.Pashkevich@imt-atlantique.fr;
Damien.Chablat@cnrs.fr.

**Abstract.** The paper focuses on the redundancy resolution in kinematic control of a new type of serial manipulator composed of multiple tensegrity segments, which are moving in a multi-obstacle environment. The general problem is decomposed into two sub-problems, which deal with collision-free path planning for the robot end-effector and collision-free motion planning for the robot body. The first of them is solved via discrete dynamic programming, the second one is worked out using quadratic programming with mixed linear equality/non-equality constraints. Efficiency of the proposed technique is confirmed by simulation.

## 1 Introduction

In robotics, kinematic control of compliant serial manipulators attracted much attention recently [1, 2, 3]. Because of their specific design including not only rigid components but also elastic elements, such manipulators allow achieving excellent flexibility and ability of shape-changing in under the environment. However, kinematic control of such manipulators is not a trivial problem, which requires redundancy resolution considering possible collisions of the robot end-effector and its body with the obstacles.

The considered manipulator is composed of multiple tensegrity segments, each of which contains two rigid triangle parts connected by a passive joint and two elastic edges with controllable preload [4]. In practice, to achieve the desired target location of the end-effector, both the end-effector and the manipulator body must avoid touching the obstacles. The latter imposes very essential constraints on the redundancy resolution, which is usually resolved via the kinematic model linearization and the classical quadratic programming with the linear equality constraint applied to the end-effector [5, 6]. In this paper, it is proposed to solve the problem sequentially, generating the collision-free path for the robot end-effector first, and collision-free motion for the robot body at the second stage. Relevant techniques are based on the discrete dynamic programming and the quadratic programming with mixed equality constraints applied to the end-effector, and the non-equality constraints applied to the manipulator segments.

## 2 Problem Statement

Let us consider a serial manipulator composed of $n$ similar segments based on dual-triangle tensegrity mechanisms, composed of rigid parts connected by passive joints whose rotation is constrained by two linear springs as shown in Fig. 1. It is assumed that the mechanism geometry is described by two triangle parameters ($a$, $b$), and the mechanism shape is defined by the central angle $q$, which is adjusted through two control inputs influencing on the lengths of the springs $L_1$ and $L_2$. More details concerning the manipulator kinematics is given in our previous paper [4], here we concentrate on the control issues and the redundancy resolution.

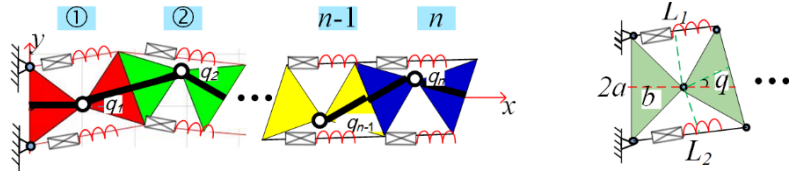

**Fig. 1.** Kinematic structure of the multi-segment serial manipulator.

For this manipulator, the direct kinematics equations can be written as follows

$$x_i = b + 2b\sum_{j=1}^{i-1}\left(\cos(\sum_{i=1}^{j} q_i)\right); \quad y_i = 2b\sum_{j=1}^{i-1}\left(\sin(\sum_{i=1}^{j} q_i)\right); \quad i=1,...,n$$
$$x_e = x_n + b\cos(\sum_{i=1}^{n} q_i); \quad y_e = y_n + b\sin(\sum_{i=1}^{n} q_i) \quad (1)$$

where $q_i$ are the joint angles, $(x_i, y_i)$ denote the position of the $i$th joint center and $(x_e, y_e)$ is the end-effector position. Corresponding Jacobians involved in the differential kinematics can be presented in the following way

$$\mathbf{J}_i = 2b \cdot \begin{bmatrix} -\sum_{k=1}^{i-1}\left(\sin\sum_{s=1}^{k} q_s\right) & -\sum_{k=2}^{i-1}\left(\sin\sum_{s=1}^{k} q_s\right) & \cdots & -\sum_{k=n}^{i-1}\left(\sin\sum_{s=1}^{k} q_s\right) \\ \sum_{k=1}^{i-1}\left(\cos\sum_{s=1}^{k} q_s\right) & \sum_{k=2}^{i-1}\left(\cos\sum_{s=1}^{k} q_s\right) & \cdots & \sum_{k=n}^{i-1}\left(\cos\sum_{s=1}^{k} q_s\right) \end{bmatrix}_{2\times n} \quad (2)$$

$$\mathbf{J}_e = \mathbf{J}_n + b \cdot \begin{bmatrix} -\sin\sum_{i=1}^{n} q_i & -\sin\sum_{i=1}^{n} q_i & \cdots & -\sin\sum_{i=1}^{n} q_i \\ \cos\sum_{i=1}^{n} q_i & \cos\sum_{i=1}^{n} q_i & \cdots & \cos\sum_{i=1}^{n} q_i \end{bmatrix}_{2 \times n} \quad (3)$$

Obviously, for $n>2$ this manipulator is kinematically redundant since the desired end-effector location can be achieved in an infinite number of ways. So, **the principle problem** considered here is how efficiently to use this kinematic redundancy in a multi-obstacle environment, i.e. to ensure the end-effector displacement to the given end-effector location $(x_e^d, y_e^d)$ with minimum joint motions $\Delta q_i$, $i=1,...,n$ while avoiding possible collisions of the manipulator body and the end-effector with the obstacles. In this paper, it is proposed to decompose these general problems into two sub-problems sequentially dealing with (i) collision-free path planning for the robot end-effector and (ii) collision-free motion planning for the robot body. More strict formalization of these problems and their solutions are presented in the following chapters.

## 3   Path Generation for the Manipulator End-effector

To find the best **collision-free path for the end-effector** let us apply the discrete dynamic programming technique allowing to generate the shortest trajectory in the obstacle-dense task space, which connects the initial and target points $\mathbf{p}^0$, $\mathbf{p}^g$ and avoids collisions with the obstacles. To apply this technique, let us discretize the task space $(x, y)$ and present it as a two-dimensional set of nodes defined in the following way

$$\mathbf{L}(i,j) = \left(x^0 + \Delta x \cdot j,\ y^0 + \Delta y \cdot i\right), \quad i=0,1,...m,\ j=0,1,...n \quad (4)$$

where $\Delta x$, $\Delta y$ are the discretization steps such that the index $j=0$ corresponds to the initial point $\mathbf{p}^0$ and the index $j=n$ corresponds to the target point $\mathbf{p}^g$. Using such presentation the desired trajectory can be presented as the sequence of the nodes

$$\mathbf{L}(i_0, 0) \to \mathbf{L}(i_1, 1) \to ... \to \mathbf{L}(i_{n-1}, n-1) \to \mathbf{L}(i_n, n) \quad (5)$$

with the purely geometric definition of the distances between the successive nodes as

$$dist\{\mathbf{L}(i,j),\ \mathbf{L}(i', j+1)\} = \sqrt{\Delta y^2 \cdot (i'-i)^2 + \Delta x^2} \quad (6)$$

To take into account possible collisions between the robot end-effector and the workspace obstacles, let us also define the binary matrix B of size $m \times n$ whose elements $\mathbf{B}(i, j) \in \{0,\ 1\}$ are equal to zero if there is no collision between the manipulator end-effector and the workspace obstacles at the node $\mathbf{L}(i, j)$, (otherwise, it is equal to one). It is worth mentioning that the above presentation neglects the robot end-effector dimensions and presents it as a point. For this reason, while computing the matrix **B** it is reasonable to modify slightly the obstacle models and increase their dimensions by the value of $\sqrt{a^2+b^2}$, where *a*, *b* are the geometric parameters of the manipulator segments (see Fig.1).

Such formalization operating with the discretized task space $\{\mathbf{L}(i,j)\}$, which includes the obstacles defined by the binary matrix **B**, allows us to present the original problem of the collision-free path planning for the manipulator end-effector as the classical shortest-path searching on the graph: *find the optimal path* (5) *on the graph connecting adjacent columns of* $\{\mathbf{L}(i,j)\}$, *which* (i) *connects the given nodes* $\mathbf{L}(i_0,0)$ *and* $\mathbf{L}(i_n,n)$, (ii) *passes through allowable nodes only* $\mathbf{B}(i,j)=0$ *and* (iii) *satisfies the optimization criterion*

$$\sum_{j=0}^{n-1} dist\{\mathbf{L}(i_j,j),\ \mathbf{L}(i_{j+1},j+1)\} \to \min_{\{i\}} \tag{7}$$

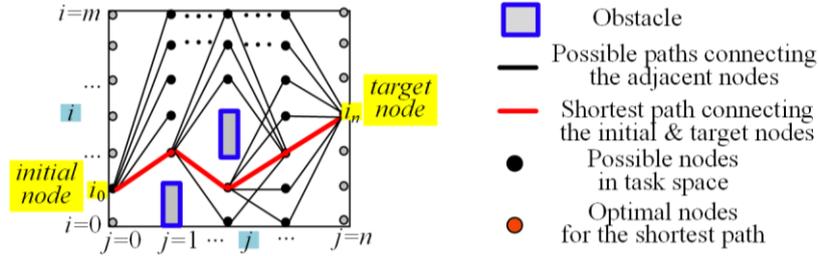

**Fig. 2.** Generation of the obstacle-free path using discrete dynamic programming

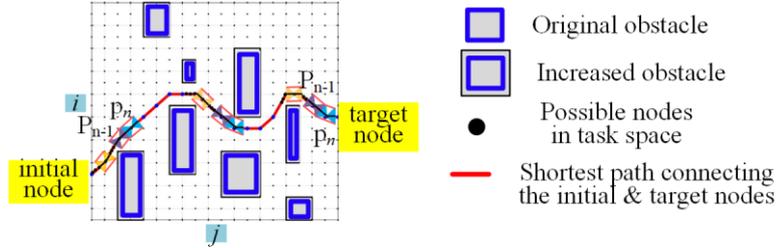

**Fig. 3.** Example of obstacle-free path generation for the robot end-effector.

It should be noted that for such presentation the desired trajectory is defined by the sequence of the row indices $\{i_0,i_1,...,i_n\}$, where both $i_0$ and $i_n$ are given (they are defined by the initial and target points). It is clear that this shortest-path problem can be solved via the discrete dynamic programming that is based on the following expression

$$d_{j+1}^*(i') = \min_i \{d_j^*(i) + dist\{\mathbf{L}(i,j),\ \mathbf{L}(i',j+1)\}\},\ \forall i'=0,1,...,m \tag{8}$$

where $d_j^*(i)$ denotes the shortest distance between the initial node $\mathbf{L}(i_0,0)$ and the node $\mathbf{L}(i,j)$ corresponding to the optimization of the lower dimension ($j \leq n$). This expression is applied sequentially starting from $j=1$ and ending with $j=n-1$, and memorizing the row indices $\{i_1^*,...,i_{n-1}^*\}$ obtained from (5) and corresponding to all intermediate optimal paths. At the final step, a single node $\mathbf{L}(i_n^*,n)$ corresponding to the desired endpoint is selected, and the desired solution is obtained through the backtracking allowing to find the remaining row indices $\{i_1^*,...,i_{n-1}^*\}$ describing the optimal path. Geometric explanation of this technique is given in Fig. 2, where the spatial location of the initial and target points corresponds to the motion "from left to right".

The efficiency of this technique has been confirmed by the simulation study. An example of obstacle-free path generation with the discretization of 20×20 is presented in Fig. 3. It should be mentioned that here, to take into account the end-effector size, the obstacles were slightly increased. As follows from this study, for such relatively rough discretization the algorithm is very fast. However, for finer discretization the computing time may increase significantly.

To overcome this difficulty, a two-step modification of the path-generation algorithm was also proposed. The basic idea of the proposed modification (leading to the algorithm speed-up) is to find first an initial solution with the rough discretization, and to improve it further using a relatively small discretization step (and applying at both steps the same numerical technique based on the discrete dynamic programming). Geometric explanation of this approach is presented in Fig. 4, where at the first step the task space is divided into several big areas $\mathbf{S}(u,v)$, $u \subset \{0,1,...m\}$, $v = \{0,1,...n\}$.

Then after applying the proposed technique, the confident areas in every column in the task space could be found, which contain the possible points for connecting the shortest path, and the corresponding trajectory could be obtained with the indices expressed as $\mathbf{S}(u_0,0) \to \mathbf{S}(u_1,1) \to ... \to \mathbf{S}(u_{n-1},n-1) \to \mathbf{S}(u_n,n)$. As the second step, it is only necessary to search for the points $\mathbf{L}(i_v,v) \in \mathbf{S}(u_v,v)$ inside of the confident areas obtained from the first step. It is clear that this approach allows us to increase significantly the computing speed.

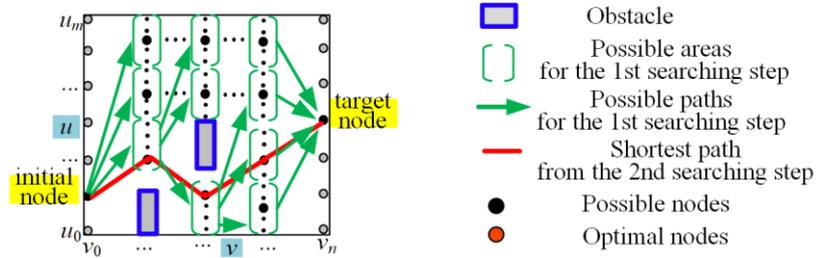

**Fig. 4.** Speed-up of the algorithm for obstacle-free path generation for the robot end-effector

## 4 Motion Generation for the Manipulator Body

To generate motions for the manipulator body it is necessary to use the best way of the manipulator redundancy, which in our case can be treated as simultaneous achievement of two goals: (i) *minimization of the joint motions for the desired end-effector location*; (ii) *ensuring safe distances between the manipulator segments and the obstacles*. The first of them can be presented as the minimization of the joint increments $\Delta \mathbf{q}$

$$\sum_{i=1}^{n} \Delta \mathbf{q}_i^T \cdot \Delta \mathbf{q} \to \min_{\Delta \mathbf{q}} \qquad (9)$$

subject to the geometric constraint

$$\Delta \mathbf{p} = \mathbf{J}_e \cdot \Delta \mathbf{q} \qquad (10)$$

arising from the desired end-effector displacement $\Delta \mathbf{p}$ computing via the kinematic Jacobian $\mathbf{J}_e$ of the manipulator end-effector. It is known that these constraint optimization problems can be easily solved analytically via the Jacobian pseudo-inverse

$$\Delta \mathbf{q} = \mathbf{J}_e^T \left( \mathbf{J}_e \mathbf{J}_e^T \right)^{-1} \Delta \mathbf{p} \qquad (11)$$

However, to take into account the second goal (collision avoidance), it is necessary to impose some additional constraints arising from the safety distances between the obstacles and the manipulator intermediate segments. It can be proved that these distances can be computed in the following way

$$d_{ij} @ dist(\mathbf{p}_i, {}^o\mathbf{p}_j) \ge d_j^0, \qquad \forall i = 1, 2, \ldots n; \quad \forall j = 1, 2, \ldots, m \qquad (12)$$

where $d_{ij}$ denotes the distance between the $i$th joint center and the $j$th obstacle, and $d_j^0$ is the allowable minimum value for the $j$th obstacle that takes into account its size (equivalent radius). In more detail, these definitions are explained in Fig. 5, where the joint axis locations are described by the points $\{\mathbf{p}_i, \forall i\}$ and the obstacles are approximated by the circles with the centers $\{{}^o\mathbf{p}_j\}$ and radiuses $\{r_j\}$.

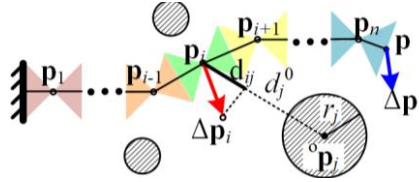

**Fig. 5.** Computing the distances d$ij$ between the robot joints and obstacles.

To present these additional constraints more conveniently, let us use the linearized expression $\Delta \mathbf{p}_i = \mathbf{J}_i \cdot \Delta \mathbf{q}$ for the manipulator joints, where $\mathbf{J}_i$ is computed from (2). Such linearization allows us to present $dist(\mathbf{p}_i, {}^o\mathbf{p}_j)$ as the projection of the displacement vector $\Delta \mathbf{p}_i$ onto the line segment connecting the points $\mathbf{p}_i$ and ${}^o\mathbf{p}_j$ (see Fig. 5), i.e.

$$d_{ij} = \mathbf{e}_{ij}^T \cdot \mathbf{J}_i \cdot \Delta \mathbf{q} \qquad (13)$$

where the unit vector $\mathbf{e}_{ij}$ is computed as $\mathbf{e}_{ij} = (\mathbf{p}_i - {}^o\mathbf{p}_j)/\|\mathbf{p}_i - {}^o\mathbf{p}_j\|$.

So finally, for the $n$ segment manipulator with $m$ different task space obstacles, the $m \times n$ collision-free constraints can be rewritten as the following way

$$\mathbf{e}_{ij}^T \cdot \mathbf{J}_i \cdot \Delta \mathbf{q} - d_j^0 \geq 0, \quad i = 1,2,...n; \quad j = 1,2,...,m \qquad (14)$$

where the safety parameter $d_j^0 = r_j + \sqrt{a^2 + b^2}$ is computed taking into account both the obstacle equivalent radius $r_j$ and the manipulator geometric parameters $a$, $b$.

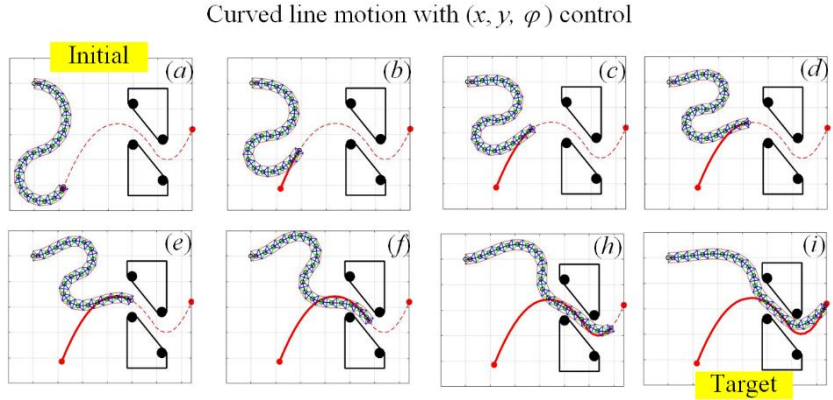

**Fig. 6.** Example of collision-free motion control for the multi-segment manipulator.

Hence, the original optimization problem with the quadratic objective (9) and linear equality constraint (10) is transformed to a more general one, which includes both the linear equality constraint (10) and a number of linear non-equality constraints (14). The main particularity of this mixed optimization problem is related to the influence of the non-equality constraints. In particular, some of them can be stronger than the other ones, leading to the situation when a limited number of non-equalities are active. In this work, it is proposed the following technique to solve this optimization problem:

1. First, try to release all non-equality constraints and find the optimal solution $\Delta \mathbf{q}^*$ of this reduced problem from (11).

2. For the obtained solution $\Delta \mathbf{q}^*$, verify all non-equality constraints (14) and find those that are violated. If no one of the constraints is violated, the final solution is obtained.
3. If some of the non-equality constraints are violated, the strongest of them is selected for each joint and transformed into the equality constraint.
4. Then the problem is solved for the extended set of equality constraints and the obtained new optimal solution $\Delta \mathbf{q}^*$ is evaluated by starting from step 2.

To find the optimal solution for the extended optimization problem at step 4, the Lagrange technique can be applied dealing with the minimization of the function

$$L(\Delta \mathbf{q}, \lambda, \mu) = \Delta \mathbf{q}^T \Delta \mathbf{q} + \lambda^T \cdot (\mathbf{J} \cdot \Delta \mathbf{q} - \Delta \mathbf{p}) + \sum_{active} \mu_{ij} \left( \mathbf{e}_{ij}^T \cdot \mathbf{J}_i \cdot \Delta \mathbf{q} - d_j^0 \right) \to \min \quad (15)$$

which leads to the following linear system

$$\Delta \mathbf{q} - \lambda^T \cdot \mathbf{J} - \mu^T \cdot \mathbf{J}_a = \mathbf{0}; \quad \mathbf{J} \cdot \Delta \mathbf{q} - \Delta \mathbf{p} = \mathbf{0}; \quad \mathbf{J}_a \cdot \Delta \mathbf{q} - \mathbf{d}_a = \mathbf{0} \quad (16)$$

where the matrix $\mathbf{J}_a$ and the vector $\mathbf{d}_a$ are composed of elements $\mathbf{e}_{ij}^T \cdot \mathbf{J}_i$ and $d_j^0$ corresponding to the active constraints, and $\lambda$ and $\mu$ are the Lagrange multipliers. It is clear that this system can be solved in a usual way via the matrix pseudo-inverse. The efficiency of the develop technique is confirmed by the simulation results presented in Fig. 6, where the manipulator end-effector must follow the curved path located inside of the narrow gap between the obstacles.

## 5  Conclusion

The paper proposes a new method of redundancy resolution in kinematic control of a new type of serial manipulator, which is moving in the multi-obstacle environment. Because of their specific design including not only rigid components but also elastic elements, such manipulators allow achieving excellent flexibility and ability of shape-changing in accordance with the environment. However, kinematic control of such manipulators requires redundancy resolution taking into account possible collisions of the robot end-effector and its body with the obstacles. To find the desired robot motion, the general problem is decomposed in two sub-problems, which deal with collision-free path planning for the robot end-effector and collision-free motion planning for the robot body. The first of them is solved via discrete dynamic programming, the second one is worked out using quadratic programming with mixed linear equality/non-equality constraints. The efficiency of the proposed technique is confirmed by simulation. In the future, this technique will be extended for the 3D manipulator with similar tensegrity segments.

## References

1. Arsenault, M., Gosselin, C.M. Kinematic, static and dynamic analysis of a planar 2-DOF tensegrity mechanism. Mechanism and Machine Theory 41, 1072–1089 (2006).
2. Furet, M., Lettl, M., Wenger, P. Kinematic Analysis of Planar Tensegrity 2-X Manipulators, in: Lenarcic, J., Parenti-Castelli, V. (Eds.), Advances in Robot Kinematics 2018. Springer International Publishing, Cham, pp. 153–160 (2019).
3. Wenger, P., Chablat, D. Kinetostatic analysis and solution classification of a class of planar tensegrity mechanisms. Robotica 37, 1214–1224 (2019).
4. Zhao, W., Pashkevich, A., Klimchik, A., Chablat, D. Stiffness Analysis of a New Tensegrity Mechanism based on Planar Dual-triangles. Presented at the 17th International Conference on Informatics in Control, Automation and Robotics, pp. 402–411. (2020)
5. Cai, B., Zhang, Y. Different-Level Redundancy-Resolution and Its Equivalent Relationship Analysis for Robot Manipulators Using Gradient-Descent and Zhang's Neural-Dynamic Methods. IEEE Transactions on Industrial Electronics 59, 3146–3155 (2012).
6. Tanaka, M., Matsuno, F. Modeling and Control of Head Raising Snake Robots by Using Kinematic Redundancy. J Intell Robot Syst 75, 53–69 (2014).